\documentclass[lettersize,journal]{IEEEtran}

\usepackage{amsmath,amsfonts}
\usepackage[ruled]{algorithm2e}
\usepackage{array}
\usepackage{textcomp}
\usepackage{stfloats}
\usepackage{url}
\usepackage{verbatim}
\usepackage{graphicx}
\usepackage{multirow}
\usepackage{bbding}
\usepackage{gensymb}
\usepackage{booktabs}
\usepackage{subfigure}
\usepackage{color}
\usepackage{xcolor}
\usepackage{threeparttable}
\usepackage{tabularx}

\ifCLASSOPTIONcompsoc
\usepackage[nocompress]{cite}
\else
\usepackage{cite}
\fi

\begin{document}
	
\title{Vision Transformer for Contrastive Clustering}

\author{Hua-Bao Ling,
	Bowen Zhu,
	Dong Huang,~\IEEEmembership{Member,~IEEE, }
	Ding-Hua Chen,\\
	Chang-Dong Wang,~\IEEEmembership{Member,~IEEE, }
	and~Jian-Huang Lai,~\IEEEmembership{Senior Member,~IEEE, }
	\IEEEcompsocitemizethanks{\IEEEcompsocthanksitem This project was supported by the NSFC (61976097, 61876193  \& 62076258) and the Natural Science Foundation of Guangdong Province (2021A1515012203). \textit{(Corresponding author: Dong Huang)}
		\IEEEcompsocthanksitem H.-B. Ling, B. Zhu, D. Huang, and D.-H. Chen are with the College of Mathematics and Informatics, South China Agricultural University, Guangzhou, China. \protect\\
		E-mail: hbling@stu.scau.edu.cn, bowenzhu0329@hotmail.com, \protect\\
		huangdonghere@gmail.com, dhchen@stu.scau.edu.cn. 
		\IEEEcompsocthanksitem C.-D. Wang and J.-H. Lai are with the School of Computer Science and Engineering,
		Sun Yat-sen University, Guangzhou, China, and also with Guangdong Key Laboratory of Information Security Technology, Guangzhou, China, and also with Key Laboratory of Machine Intelligence and Advanced Computing, Ministry of Education, China.\protect\\
		E-mail: changdongwang@hotmail.com, stsljh@mail.sysu.edu.cn.}
	\thanks{H.-B. Ling and B. Zhu contribute equally to this work.}
}

\maketitle

\begin{abstract}
Vision Transformer (ViT) has shown its advantages over the convolutional neural network (CNN) with its ability to capture global long-range dependencies for visual representation learning. Besides ViT, contrastive learning is another popular research topic recently. While previous contrastive learning works are mostly based on CNNs, some recent studies have attempted to combine ViT and contrastive learning for enhanced self-supervised learning. Despite the considerable progress, these combinations of ViT and contrastive learning mostly focus on the instance-level contrastiveness, which often overlook the global contrastiveness and also lack the ability to directly learn the clustering result (e.g., for images). In view of this, this paper presents a novel deep clustering approach termed \textit{V}ision \textit{T}ransformer for \textit{C}ontrastive \textit{C}lustering (VTCC), which for the first time, to our knowledge, unifies the Transformer and the contrastive learning for the image clustering task. Specifically, with two random augmentations performed on each image, we utilize a ViT encoder with two weight-sharing views as the backbone. To remedy the potential instability of the ViT, we incorporate a convolutional stem to split each augmented sample into a sequence of patches, which uses multiple stacked small convolutions instead of a big convolution in the patch projection layer. By learning the feature representations for the sequences of patches via the backbone, an instance projector and a cluster projector are further utilized to perform the instance-level contrastive learning and the global clustering structure learning, respectively. Experiments on eight image datasets demonstrate the stability (during the training-from-scratch) and the superiority (in clustering performance) of our VTCC approach over the state-of-the-art. The code is available at \url{https://github.com/JackKoLing/VTCC}.
\end{abstract}



\section{Introduction}\label{sec:introduction}
\IEEEPARstart{V}{ision}  Transformer (ViT) \cite{DosovitskiyB0WZ21}, as a promising alternative network architecture to the convolutional neural network (CNN) \cite{he2016deep}, has received increasing attention recently. Although the CNN can effectively capture the local information, there is still a practical need to model the global long-range dependencies between image elements in many computer vision tasks. Inspired by the significant success of the Transformer architecture that adopts the multi-head self-attention (MHSA) mechanism \cite{vaswani2017attention} in natural language processing (NLP), the ViT has recently emerged as a pioneering technique to apply the Transformer architecture in computer vision \cite{DosovitskiyB0WZ21}. Typically, in ViT, an input image is first split into a sequence of patches by a non-overlapping $p \times p$ convolutional operation with stride-$p$ (e.g., with $p=16$). Then these patches are fed into a standard Transformer encoder architecture \cite{vaswani2017attention} for representation learning. With this simple yet powerful backbone architecture, many Transformer-based models have been developed and some have become the predominant architectures for a variety of vision tasks, such as image classification \cite{DosovitskiyB0WZ21}, object detection \cite{carion2020end}, and semantic segmentation \cite{strudel2021segmenter}.

In the meantime, the self-supervised learning (SSL) \cite{chen2020simple,he2020momentum} has gained remarkable progress in the past few years, which aims to learn visual representation from unlabeled data in a self-supervised manner. There are three mainstream categories in SSL \cite{liu2021self}, which are generative \cite{van2016conditional}, contrastive \cite{chen2020simple,he2020momentum}, and generative-contrastive \cite{radford2015unsupervised} architectures, respectively. Among them, the contrastive learning has recently made some breakthroughs, which aims to ``learn to compare" via an information noise contrastive estimation (InfoNCE) \cite{oord2018representation} loss function. As a representative work, MoCo \cite{he2020momentum} utilizes the momentum contrastive learning with two encoders and employs a queue to save negative samples. Alternatively, SimCLR \cite{chen2020simple} demonstrates the importance of constructing the samples pairs by various data augmentations, and can be trained with a large batch size instead of the momentum contrast in MoCo \cite{he2020momentum}.

Rapid progress has been made in both the Transformer-based architectures and the contrastive learning architectures. However, these two types of architectures are mostly developed independently, and almost all contrastive learning backbones are based on the CNN. A critical question arises as to whether the ViT can provide better expressiveness for contrastive learning than the CNN. Recently, some research works have attempted to exploit the advantages of both of them. For example, Chen et al. \cite{chen2021empirical} studied the instability issue on training the self-supervised ViT. Caron et al. \cite{caron2021emerging} explored some emerging properties of the self-supervised ViT.
In contrastive learning, a recent development is its combination with the image clustering task. Typically, Li et al. \cite{li2021contrastive} utilized the contrastive learning at the instance-level and the cluster-level for simultaneous representation learning and image clustering, where the CNN is adopted as the backbone. As the CNN is more focused on local information, it is intuitive that the modeling of the global dependencies via Transformer may provide rich information to enhance the image clustering performance, especially for complex images. Yet surprisingly, it remains an unaddressed problem how to effectively leverage the Transformer (or ViT) in collaboration with the contrastive learning for joint representation learning and image clustering.

Notably, this paper for the first time, to the best of our knowledge, unifies the ViT and the contrastive learning for the image clustering task. To make this unification possible and practical, some critical challenges should be handled.
Different from CNN \cite{he2016deep}, the training recipes for ViT are often sensitive, especially in the self-supervised scenarios \cite{chen2021empirical}, which involve the choice of hyper-parameters, the number of encoder blocks, and so forth. As a consequence, the training of ViT usually needs a large dataset, and the instability issue is often encountered \cite{caron2021emerging,liu2021efficient,xiao2021early}. Recently, some studies suggest that this issue may arise from the path projection layer in the standard ViT. Regarding this, Chen et al. \cite{chen2021empirical} alleviated the instability issue by freezing the patchify stem instead of using the random initialization patch projection layer. Xiao et al. \cite{xiao2021early} replaced the origin patchify stem with a convolutional stem, which is implemented by a small number of stacked stride-$2$ $3\times3$ convolutions in the patch projection layer. In this work, we observe that the use of multiple stacked small convolutions instead of a big convolution in the patch projection layer, which to some extent resembles the typical design of CNNs \cite{he2016deep,tan2019efficientnet},  can significantly benefit the contrastive image clustering with Transformer, even for some relative small datasets.

\begin{figure}[!t]
	\begin{center}
		{\includegraphics[width=0.99\columnwidth]{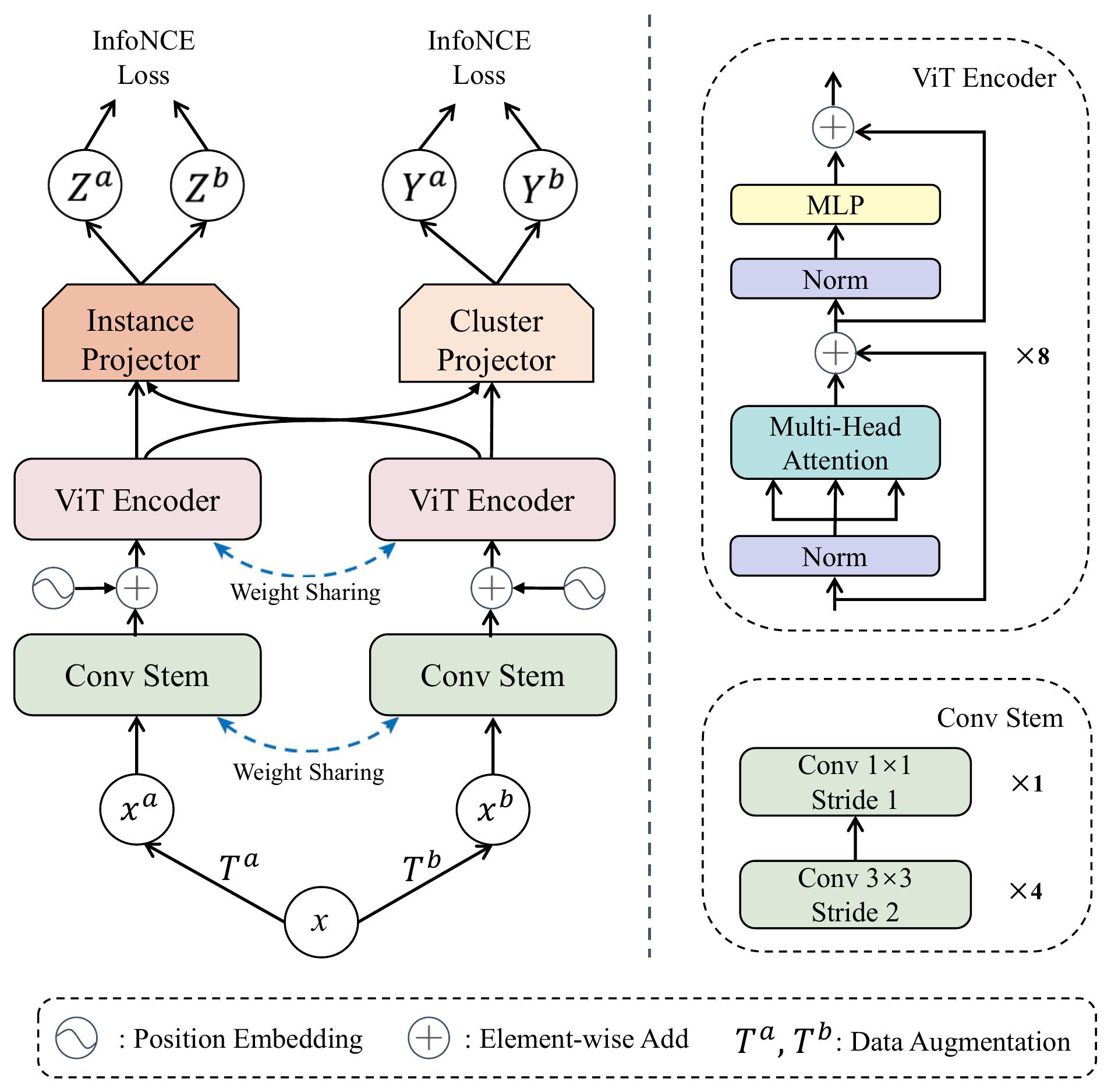}}
		\caption{\textbf{Vision Transformer for Contrastive Clustering (VTCC)}. }
		\label{fig}
	\end{center}
\end{figure}

Specifically, we propose a novel deep clustering approach termed \textbf{V}ision \textbf{T}ransformer for \textbf{C}ontrastive \textbf{C}lustering (VTCC) (as illustrated in Fig.~\ref{fig}). To remedy the potential instability, a convolutional stem layer with multiple stacked small-size convolutions is incorporated to split each augmented sample image into a sequence of patches. With two random augmentations performed on each image to obtain two augmented samples, we feed the sequences of patch vectors of the augmented samples to the backbone, which encompasses two weight-sharing views in a standard Transformer encoder architecture. With the representations of the augmented samples learned by the self-attention mechanism, the contrastive learning is enforced via an instance projector and a cluster projector, which explore the instance contrastiveness and the global clustering structure, respectively. Note that the Transformer backbone with convolutional stem and the two contrastive projectors in VTCC are jointly trained in an end-to-end manner. Extensive experiments are conducted on eight challenging image datasets, which confirm the benefits brought in by the Transformer, and demonstrate the stability (during the training-from-scratch) and the superior clustering performance of VTCC in comparison with the state-of-the-art.

The main contributions of this work are summarized below.
\begin{itemize}
	\item This paper makes the first attempt to effectively bridge the gap between the Transformer and the contrastive learning for the deep image clustering task, with the global dependencies, the instance contrastiveness, and the clustering structure learning jointly modeled.
	\item This paper incorporates a convolutional stem with multiple stacked small convolutions in the patch projection layer, which enhances the stability during the ViT training from scratch for image clustering.
	\item This paper presents an end-to-end deep clustering approach termed VTCC with ViT and contrastive learning jointly modeled. Experimental results on eight real-world image datasets demonstrate the promising capability of Transformer for the clustering of complex images and the advantages of VTCC over the state-of-the-art deep clustering approaches.
\end{itemize}

The remainder of the paper is organized as follows. The proposed VTCC approach is described in Section~\ref{sec:proposed framework}. The experimental results are provided in Section~\ref{sec:experiments}, followed by the conclusion in Section~\ref{sec:conclusion}.

\section{Proposed Framework}
\label{sec:proposed framework}

In this section, we describe the details of our VTCC approach. First, an overview of the network architecture is given in Section~\ref{sec:network architecture}. Then the ViT backbone is presented in Section~\ref{sec:backbone with vision transformer}. Finally, the overall loss function of VTCC is provided in Section~\ref{sec:loss function}.

\subsection{Network Architecture}
\label{sec:network architecture}

The architecture of VTCC is illustrated in Fig.~\ref{fig}, which consists of three main components, namely, the ViT backbone with two weight-sharing views to extract feature representations for augmented samples, the instance-level contrastive learning module to maximize the similarity of positive pairs, and the global clustering structure learning module that obtains the soft label prediction for image clustering.

Specifically, for each image in a mini-batch of $N$ images, two data augmentations $T^a$ and $T^b$ are randomly selected with a certain probability so as to obtain $2$ augmented samples for this image, thus obtaining $2\cdot N$ augmented samples for this mini-batch. Then each of the augmented samples is split into a sequence of patches by the convolutional stem with multiple stacked small convolutions in the patch projection layer \cite{xiao2021early}. With each sample split into a sequence of patches and embedded by linear projection, a standard Transformer encoder is utilized for learning the feature representations, which are then fed to two projectors, i.e., the instance projector and the cluster projector, for instance-level contrastive learning and global clustering structure learning, respectively. The overall network is trained by two contrastive loss functions in an end-to-end manner, through which the clustering result can therefore be learned. Additionally, the pseudo-code of VTCC is presented in Algorithm~\ref{alg}.

\definecolor{c1}{HTML}{006E5F}
\definecolor{c5}{HTML}{ED9F54}

\begin{algorithm}[h]
	\label{alg}
	\caption{VTCC: A PyTorch-like Pseudo Code}
	\LinesNumbered
	\textcolor{c1}{\# $f(\cdot)$ : Vision Transformer}\\
	\textcolor{c1}{\# $g_I(\cdot)$ : Instance projector}\\
	\textcolor{c1}{\# $g_C(\cdot)$ : Cluster projector}\\
	\textcolor{c1}{\# $InfoNCE(\cdot,\cdot)$ : InfoNCE loss}\\
	\textcolor{c1}{\# $InsLoss(\cdot,\cdot)$ : Instance-level contrastive loss}\\
	\textcolor{c1}{\# $CluLoss(\cdot,\cdot)$ : Global clustering contrastive loss}\\
	\textcolor{c1}{\# $T^a,\; T^b$ : Augmentations selected from family $T$}\\
	\textcolor{c5}{for} x \textcolor{c5}{in} loader: \textcolor{c1}{ \# Load a mini-batch $x$ with $N$ samples}\\
	\qquad \textcolor{c1}{ \# Obtain augmented views of $x$}\\
	\qquad $x^a, \; x^b \;=\; T^a(x),\; T^b(x)$\\
	
	\qquad \textcolor{c1}{ \# Extract feature representations}\\
	\qquad $h^a, \; h^b \;=\; f(x^a),\; f(x^b)$\\
	\qquad \textcolor{c1}{ \# Map representations to subspace via $g_I$}\\
	\qquad $Z^a, \; Z^b \;=\; g_I(h^a),\; g_I(h^b)$\\
	
	\qquad \textcolor{c1}{ \# Map representations to subspace via $g_C$}\\
	\qquad $Y^a, \; Y^b \;=\; g_C(h^a),\; g_C(h^b)$\\\quad\\
	
	\qquad \textcolor{c1}{ \# Loss calculation}\\
	\qquad $loss = InsLoss(Z^a, Z^b) + CluLoss(Y^a, Y^b)$\\
	\qquad $loss.backward()$\\ \quad \\
	
	\qquad $update(f)$
	\textcolor{c1}{ \# Optimize the parameters of $f$}\\\quad\\
	
	\textcolor{c5}{def} $InsLoss(Z^a, Z^b)$:\\\qquad
	\textcolor{c5}{return}\quad {$(\text{InfoNCE}(Z^a) + \text{InfoNCE}(Z^b))/2$}\\
	
	\textcolor{c5}{def} $CluLoss(Y^a, Y^b)$:\\\qquad
	\textcolor{c5}{return}\quad {$(\text{InfoNCE}(Y^a) + \text{InfoNCE}(Y^b))/2$}\\
	
\end{algorithm}

\subsection{Backbone with Vision Transformer}
\label{sec:backbone with vision transformer}

Unlike the CNNs which are performed on the images directly, the Transformer encoder requires a sequence of vectors (or patches) as input. In VTCC, we utilize a convolutional stem with multiple stacked small-size convolutions to split each augmented sample into a sequence of patches, which are fed into a ViT \cite{DosovitskiyB0WZ21} backbone. In the following, we describe the convolutional stem and the Transformer encoder in Sections~\ref{sec:convolutional stem} and \ref{sec:encoder}, respectively.

\subsubsection{Convolutional Stem}
\label{sec:convolutional stem}

To convert a 2D image to a 1D sequence of feature embedding, a conventional strategy adopted in ViT is to apply the $p \times p$ convolution operation with a stride $p$ to the input image (typically with $p=16$), which is referred to as \textit{the patchify stem}. Thus, it can directly capture a lot of non-overlapping patches with the $p \times p$ size, and then flatten these patches to form the input sequence of vectors for the Transformer encoder. However, on the one hand, this type of convolution operation with a large convolution kernel and a large stride runs counter to the classical convolution setting in canonical neural networks \cite{he2016deep} which usually adopt some small-size (e.g., $3 \times 3$) convolution kernels. On the other hand, the in-patch structure information often focus on relatively small receptive field, which overlooks the rich information of local associations (e.g., the spatial relationship between patches) and to some extent makes the patchify stem not sufficient with the simple linear projections.

Regarding this issue, Xiao et al. \cite{xiao2021early} proposed a convolutional stem which replaces the patchify stem in the original ViT model by a small number of stacked stride-$2$ $3 \times 3$ convolutions. We empirically show that using the convolutional stem on the ViT patch embedding layer is particularly beneficial for our image clustering task (as evaluated in Section~\ref{sec:ablation study}). Specifically, to obtain the 1D image sequence, we first use four blocks in the convolutional stem, where each block includes a standard convolution operation with a stride-2 $3\times 3$ convolution kernel, followed by the batch normalization and a ReLU activation function (as shown in Fig.~\ref{fig}). Since the Transformer encoder requires the 1D image sequence as input, we use a single $1\times 1$ convolution to match the input dimension of the Transformer encoder. In practice, compared with the large convolution, using multiple small convolutions can capture the fine-grained local features which are more conducive to the representation learning process and further contribute to better stability and clustering performance.

\subsubsection{Transformer Encoder}
\label{sec:encoder}

In the literature, ViT models with various scales have been developed. In this paper, we adopt the ViT-Small \cite{touvron2021training} as the backbone which provides effective representation learning capability while maintaining high efficiency.
Specifically, based on the input sequence of patches embedding, each encoder layer is built upon a standard Transformer architecture that consists of a multi-head self-attention and a MLP block. In addition, a layer-norm operation is adopted \textit{before} each block, and the residual connection is incorporated \textit{after} each block. Here, the Transformer encoder applies the self-attention layers to model global relations between input embeddings, which has an advantage over the CNNs whose receptive field of convolutional kernels are limited locally. However, the Transformer architecture is permutation-invariant theoretically, which may neglect the rich information of relative positions. Therefore, we also incorporate the position encodings \cite{vaswani2017attention} to the input sequence of each self-attention layer.

\subsection{Loss Function}
\label{sec:loss function}
To simultaneously perform the instance-level and cluster-level self-supervision, two independent projectors are utilized to map the feature representations extracted by the backbone to different subspaces (i.e., the instance subspace and the cluster subspace). By means of these two projectors, the instance-level contrastive learning and the cluster-level contrastive learning are respectively enforced.

\subsubsection{Instance-level Contrastive Learning}
\label{sec:instance-level contrastive learning loss}

The purpose of the instance-level contrastive learning is to maximize the similarity between two different augmented samples of the same input image by pulling closer the distance between a positive pair while pushing away the representations of the negative pairs in the instance subspace. Let $x^a_i$ and $x^b_i$ denote the two augmented samples for the input image $x_i$, and $h^a_i$ and $h^b_i$ denote the feature representations of $x^a_i$ and $x^b_i$, respectively, extracted by the ViT encoder. Since directly using the feature representations $h^a_i$ and $h^b_i$ for similarity calculation may lead to the loss of information \cite{chen2020simple}, we utilize an instance projector with a three-layer MLP, which is denoted as $g_I$ and maps the representations $h^a_i$ and $h^b_i$ to a low-dimensional subspace, leading to $Z^a_i=g_I(h^a_i)$ and $Z^b_i=g_I(h^b_i)$, respectively. Then, the cosine similarity is used to measure the pair-wise similarity, that is
\begin{equation}
	s(\alpha, \beta) = \frac{\alpha^{\top} \beta}{\|\alpha\|\cdot\|\beta\|},
	\label{equ:cosine_sim}
\end{equation}
where $\alpha$ and $\beta$ are two feature vectors with the same dimension.

Let $I$ denote the instance-level representations set, which contains all the instance-level augmented representations in a mini-batch of $N$ images, that is, $I =  \{Z_1^a,\dots,Z_N^a,Z_1^b,\dots,Z_N^b\}$, where $Z_i^a$ (or $Z_i^b$) is the instance-level representation of sample $x_i^a$ (or $x_i^b$, respectively). Thus we can obtain $2\cdot N-1$ sample pairs for each augmented sample (say, $Z^a$), that is, $\{Z^a_i,Z^k_j\}$ with $Z^a_i\neq Z^k_j$, $k\in \{a,b\}$ and $j \in [1,N]$. Among these pairs, only the pair $\{Z_i^a,Z_i^b\}$ is the  positive pair for $Z_i^a$, while the rest are negative pairs. Then the InfoNCE loss \cite{oord2018representation} for calculating the instance-level contrastiveness is defined as
\begin{equation}
	\begin{aligned}
		&l^a_i =-log\frac{\exp(s(Z^a_i,Z^+)/\tau_I)}{\sum\limits_{Z^-\in I}
			\exp(s(Z^a_i,Z^-)/\tau_I)}
		\label{equ:loss_ins}
	\end{aligned}
\end{equation}
where $\tau_I$ is the instance-level temperature parameter, $Z^+$ and $Z^-$ are the positive and negative samples of the current sample, respectively. Thus for the image $x_i$, the instance-level loss is calculated as
\begin{equation}
	l_i = l^a_i + l^b_i,
\end{equation}
where $l^a_i$ is the instance-level contrastive loss of $x_i$ in the first view (and $l^b_i$ for the second view).
Finally, the instance-level contrastive loss for a mini-batch of $N$ input images is defined as \cite{chen2020simple}
\begin{equation}
	L_{ins} = \frac{\sum_{i=1}^N{l_i}}{2N}.
	\label{equ:ins_loss}
\end{equation}

\subsubsection{Global Clustering Structure Learning}
\label{sec:global clustering structure learning loss}

Besides the instance-level contrastive loss, we proceed to define the cluster-level contrastive loss for global clustering structure learning. It aims to maximize the similarity of the positive cluster pairs while minimizing the similarity of the negative cluster pairs. A cluster projector $g_C$ with a three-layer MLP and an extra softmax layer is incorporated to map the feature representations to the low-dimensional subspace, leading to $Y^a_i=g_C(h^a_i)$ and $Y^b_i=g_C(h^b_i)$. For a mini-batch of $N$ input images, we can obtain the cluster distribution probability matrices $Y^a \in \mathbb{R}^{N \times K}$ and $Y^b \in \mathbb{R}^{N \times K}$ for the two augmented views by stacking the learned representations of the $N$ samples, where $K$ is the number of clusters and the $i$-th column (i.e., the representation of $i$-th cluster, denoted as $\hat{Y}^k_i$ with $k\in\{a,b\}$) of these two matrices indicates the probability of allocating each sample to the $i$-th cluster.

Let $C$ denote the cluster-level representation set which contains the representations of all clusters (in the two augmented views), that is, $C = \{\hat{Y}_1^a,\dots,\hat{Y}_K^a,\hat{Y}_1^b,\dots,\hat{Y}_K^b\}$.
We take the cluster representations from $C$ to construct cluster pairs. Formally, for each cluster, say, $\hat{Y}^a_i$, we can obtain $2\cdot K-1$ cluster pairs, that is, $\{\hat{Y}^a_i,\hat{Y}^k_j\}$ with $\hat{Y}^k_j\neq \hat{Y}^a_i$, $k\in \{a,b\}$ and $j \in [1,K]$. 
For the cluster $\hat{Y_i^a}$, only the pair $\{\hat{Y_i^a},\hat{Y_i^b}\}$ is a positive cluster pair, while the rest are negative pairs. By using the cosine similarity to measure the similarity between the cluster representations, the cluster-level contrastive loss for $\hat{Y}^a_i$ is defined as
\begin{equation}
	\begin{aligned}
		&\hat{l^a_i} =-log\frac{\exp(s(\hat{Y}^a_i,Y^+)/\tau_C)}{\sum\limits_{Y^-\in C}
			\exp(s(\hat{Y}^a_i,Y^-)/\tau_C)}
		\label{equ:loss_clu}
	\end{aligned}
\end{equation}
where $\tau_C$ is the cluster-level temperature parameter, $Y^+$ and $Y^-$ are the positive and negative clusters of the current cluster, respectively. The contrastive loss for the both augmented views of the $i$-th cluster is defined as
\begin{equation}
	\hat{l_i} = \hat{l_i^a}+ \hat{l_i^b},
\end{equation}
where $\hat{l_i^a}$ is the cluster-level contrastive loss of the $i$-th cluster in the first view (and $\hat{l_i^b}$ in the second view). Then the cluster-level contrastive loss for the $K$ clusters is defined as \cite{li2021contrastive}
\begin{equation}
	L_{clu} = \frac{1}{2K}\sum_{i=1}^K\hat{l_i}-Entropy(p(Y^a))-Entropy(p(Y^b)).
	\label{equ:clu_loss}
\end{equation}

To avoid the trivial solution that assigns most samples to a single cluster, the entropy term is incorporated into the loss function, which is defined as
\begin{equation}
	Entropy(p(Y^k)) = -\sum_{i=1}^{K}[p(\hat{Y}_i^{k})logp(\hat{Y}_i^{k})],
	\label{equ:H}
\end{equation}

\begin{equation}
	\begin{array}{ll}
		p(\hat{Y}_i^{k}) = \frac{{\sum^N_{j=1}}y^k_{ji}}{{\|\hat{Y}_i^{k}\|}_1},\;
		\text{for}\; k \in \{a,b\},
		\label{equ:P}
	\end{array}
\end{equation}
where $y^a_{ji}$ (or $y^b_{ji}$) denotes the entry in the $j$-th row and $i$-th column of the representation matrix $Y^a$ (or $Y^b$, respectively).

\subsubsection{Overall Loss Function}
\label{sec:overall loss function}

By minimizing the losses~(\ref{equ:ins_loss}) and~(\ref{equ:clu_loss}), the network can be trained by maximizing the similarity between the representations of the positive pairs, while minimizing the similarity between the representations of the negative pairs, at the instance-level and the cluster-level, respectively.

In VTCC, we jointly optimize the losses $L_{ins}$ and $L_{clu}$ to enforce the instance-level and cluster-level contrastive learning and obtain the clustering result. The overall loss function is defined as
\begin{equation}
	L_{total} = L_{ins} + L_{clu}.
	\label{equ:total}
\end{equation}

\begin{table}[!t]
	\caption{The image datasets used in our experiments.}\vskip -0.08in
	\label{table-dataset}
	\centering
	\renewcommand{\arraystretch}{1.2}
	\setlength{\tabcolsep}{6.5mm}{
		\begin{tabular}{l c c}
			\hline
			Dataset   & \#Samples & \#Classes \\
			\hline
			RSOD \cite{long2017accurate}  & 976  & 4 \\
			UC-Merced \cite{yang2010bag}  & 2,100  & 21 \\
			SIRI-WHU \cite{zhao2015dirichlet}  & 2,400  & 12 \\
			AID \cite{xia2017aid}  & 10,000  & 30 \\
			D0 \cite{xie2018multi}      & 4,508 & 40 \\
			DTD \cite{cimpoi2014describing}      & 5,640 & 47 \\
			Chaoyang \cite{zhu2021hard}     & 6,160 & 4 \\
			CIFAR-100 \cite{krizhevsky2009learning}      & 60,000 & 20 \\
			\hline
	\end{tabular}}
\end{table}

\begin{figure}[!t] \vskip 0.1 in
	\centering
	\subfigure[RSOD]{
		\includegraphics[width=0.23\textwidth]{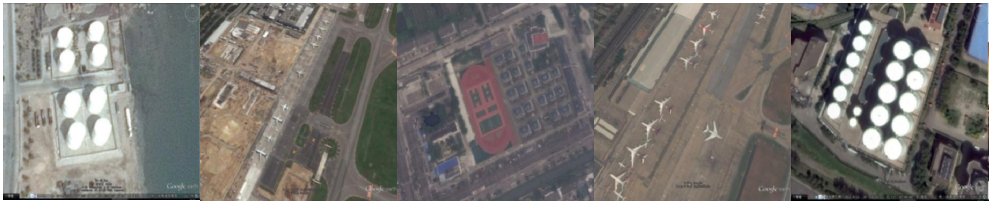}}
	\subfigure[UC-Merced]{
		\includegraphics[width=0.23\textwidth]{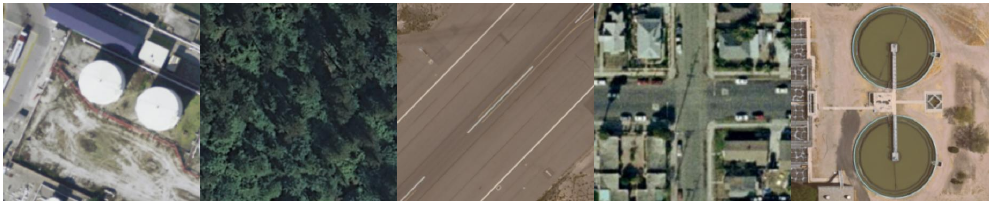}}\vskip -0.015in
	\subfigure[SIRI-WHU]{
		\includegraphics[width=0.23\textwidth]{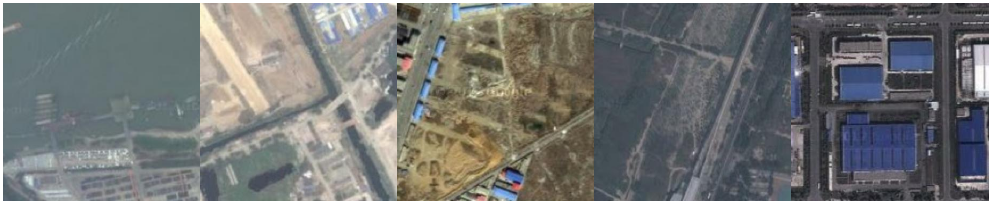}}
	\subfigure[AID]{
		\includegraphics[width=0.23\textwidth]{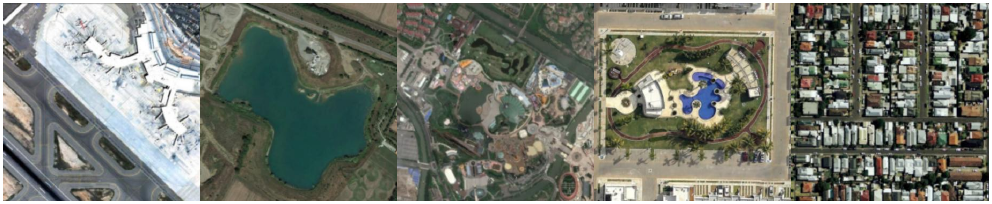}}\vskip -0.015in
	\subfigure[D0]{
		\includegraphics[width=0.23\textwidth]{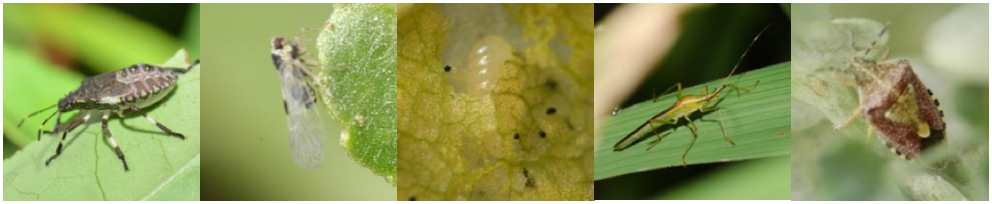}}
	\subfigure[DTD]{
		\includegraphics[width=0.23\textwidth]{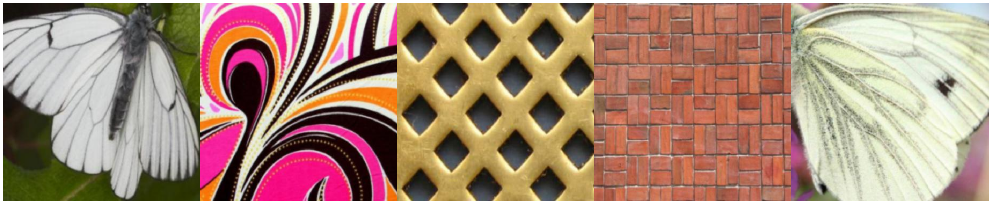}}\vskip -0.015in
	\subfigure[Chaoyang]{
		\includegraphics[width=0.23\textwidth]{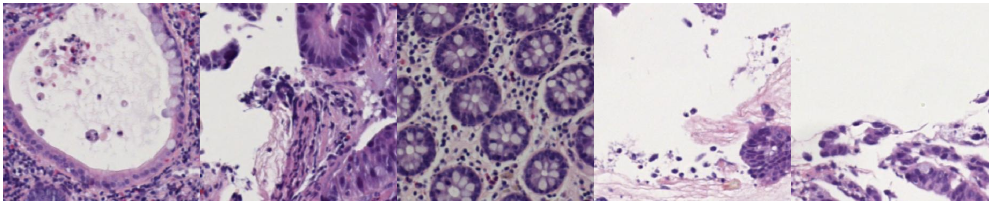}}
	\subfigure[CIFAR-100]{
		\includegraphics[width=0.23\textwidth]{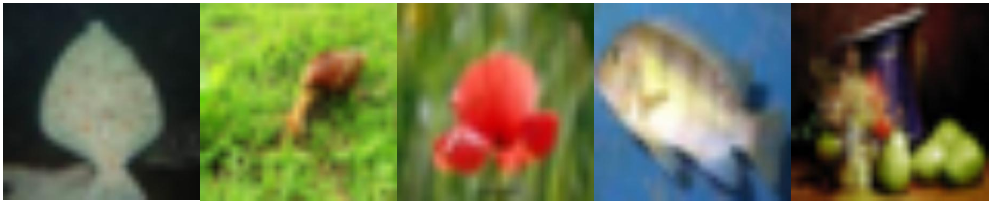}}
	\caption{Some examples of the eight real-world image datasets used for evaluation, including four remote sensing datasets \cite{yang2010bag,long2017accurate,zhao2015dirichlet,xia2017aid}, a crop pest dataset \cite{xie2018multi}, a texture dataset \cite{cimpoi2014describing}, a medical dataset \cite{zhu2021hard}, and an object image dataset \cite{krizhevsky2009learning}.}
	\label{fig:dataset_samples}
\end{figure}

\section{Experiments}
\label{sec:experiments}

In this section, we conduct experiments on eight real-world image datasets to evaluate the proposed VTCC approach against a variety of non-deep and deep clustering approaches. 

\subsection{Implementation Details}
\label{sec:implementation details}

In this work, we adopt the setting of data augmentations in BYOL \cite{grill2020bootstrap} and use the Adam optimizer with an initial learning rate of 0.0003. A lightweight version of ViT, i.e., ViT-Small \cite{touvron2021training}, is utilized as our backbone with a dimension size of 384 and 8 blocks of encoder layers. The input image size is set to $224 \times 224$. We leverage a convolutional stem with four small $3 \times 3$ convolutions instead of a large $16 \times 16$ convolution, followed by $1 \times 1$ convolution. The instance projector and the cluster projector each uses a three-layer MLP, following the projector network of MoCov3 \cite{chen2021empirical}. Besides, we set the output dimension of the instance projector to 128, which can preserve sufficient information for the instance-level contrastive learning. As for the cluster projector, the output dimension is set to the number of clusters, i.e., the true number of classes on each dataset. The temperature parameters $\tau_I$ and $\tau_C$ are set to 0.5 and 1.0, respectively. We train the VTCC architecture from scratch for 1,000 epochs with a batch size of 128. All experiments are conducted on a single NVIDIA RTX 3090 GPU and CUDA 11.0.

\subsection{Datasets and Evaluation Metrics}
\label{sec:datasets and evaluation metrics}

In our experiments, eight real-world image datasets are used for evaluation, which are described as follows:
\begin{itemize}
    \item \textbf{RSOD} \cite{long2017accurate} is a classical dataset for the remote sensing task, which consists of 976 images with 4 different classes, including aircraft, oil tank, overpass, and playground.
    \item \textbf{UC-Merced} \cite{yang2010bag} is a collection of 2,100 images with 21 different scenes and 100 images per category.
    \item \textbf{SIRI-WHU} \cite{zhao2015dirichlet} is another remote sensing image dataset, which contains a total of 2,400 images in 12 categories, where each category includes 200 images.
    \item \textbf{AID} \cite{xia2017aid} is a large remote sensing dataset, including 30 categories of scene images and a total of 10,000 images.
    \item \textbf{D0} \cite{xie2018multi} consists of 4,508 images with 40 pest classes, which were collected under real-world conditions in a variety of fields.
    \item \textbf{DTD} \cite{cimpoi2014describing} is a complex texture dataset, which consists of 5,640 images with 47 different classes.
    \item \textbf{Chaoyang} \cite{zhu2021hard} is a medical dataset with 6,160 images and 4 classes of the colon slides, collected from Chaoyang hospital.
    \item \textbf{CIFAR-100} \cite{krizhevsky2009learning} is a widely-used object image dataset. We use its 20 super-classes as the ground-truth.
\end{itemize}

For clarity, we summarize the statistics of the eight image dataset in Table~\ref{table-dataset} and illustrate some sample images of them in Fig.~\ref{fig:dataset_samples}.

Following the standard evaluation protocol for the image clustering task, we adopt three widely-used evaluation metrics in the experiments, including the normalized mutual information (NMI) \cite{strehl2002cluster}, the accuracy (ACC) \cite{Huang2020}, and the adjusted Rand index (ARI) \cite{Huang2021}. Note that higher values of these metrics indicate better clustering results.

\begin{table*}[!t]\vskip 0.1in
	\centering
	\caption{The \textbf{NMI} scores of different image clustering methods on eight datasets.}\vskip -0.08in
	\label{tab-NMI}
\begin{threeparttable}
\begin{tabular}{m{2cm}<{}|m{1.5cm}<{\centering}m{1.5cm}<{\centering}m{1.5cm}<{\centering}m{1.5cm}<{\centering}m{1.5cm}<{\centering}m{1.5cm}<{\centering}m{1.5cm}<{\centering}m{1.5cm}<{\centering}}
		\toprule
		Dataset &RSOD  &UC-Merced &SIRI-WHU &AID   &D0    &DTD   &Chaoyang &CIFAR-100\\
		\midrule
		$K$-means {\cite{macqueen1967some}}	&0.162	&0.204	&0.145	&0.209	&0.299	&0.119	&0.024	&0.083\\
		SC {\cite{zelnik2005self}}	&0.146	&0.211	&0.161	&0.189	&0.305	&0.118	&0.022	&0.090\\
		AC {\cite{gowda1978agglomerative}}	&0.168	&0.214	&0.166	&0.204	&0.319	&0.125	&0.026	&0.098\\
		NMF {\cite{cai2009locality}}	&0.176	&0.202	&0.245	&0.193	&0.255	&0.127	&0.018	&0.076\\
		PCA {\cite{martinez01_pca}}	&0.163	&0.206	&0.164	&0.216	&0.308	&0.124	&0.024	&0.084\\
		BIRCH {\cite{Zhang1996}}	&0.148	&0.225	&0.162	&0.204	&0.315	&0.123	&0.026	&-\\
		GMM	{\cite{fraley03}} &0.160	&0.198	&0.160	&0.205	&0.289	&0.120	&0.024	&0.084\\
		\midrule
		DEC	{\cite{xie2016unsupervised}} &0.296	&0.120	&0.183	&0.217	&0.328	&0.128	&0.001	&0.101\\
		IDEC {\cite{guo2017improved}}	&0.209	&0.119	&0.178	&0.207	&0.309	&0.128	&0.001	&0.103\\
		ASPC-DA {\cite{guo2019adaptive}}	&0.054	&0.137	&0.103	&0.060	&0.153	&0.084	&0.026	&-\\
		IDFD {\cite{tao2021clustering}}	&0.391	&0.572	&0.540	&0.696	&0.663	&0.410	&0.309	&0.428\\
		CC {\cite{li2021contrastive}}	&0.457	&0.609	&0.603	&0.752	&0.693	&0.480	&0.365	&0.424\\
		VTCC (Ours)	&\textbf{0.611}	&\textbf{0.658}	&\textbf{0.693}	&\textbf{0.794}	&\textbf{0.753}	&\textbf{0.487}	&\textbf{0.373}	&\textbf{0.432}\\
		\bottomrule
	\end{tabular}
\end{threeparttable}
\end{table*}

\begin{table*}[!t]\vskip 0.05in
	\centering
	\caption{The \textbf{ACC} scores of different image clustering methods on eight datasets.}\vskip -0.08in
	\label{tab-ACC}
\begin{threeparttable} \begin{tabular}{m{2cm}<{}|m{1.5cm}<{\centering}m{1.5cm}<{\centering}m{1.5cm}<{\centering}m{1.5cm}<{\centering}m{1.5cm}<{\centering}m{1.5cm}<{\centering}m{1.5cm}<{\centering}m{1.5cm}<{\centering}}
		\toprule
		Dataset &RSOD  &UC-Merced &SIRI-WHU &AID   &D0    &DTD   &Chaoyang &CIFAR-100\\
		\midrule
		$K$-means \cite{macqueen1967some}	&0.388	&0.200	&0.229	&0.163	&0.204	&0.090	&0.320	&0.137\\
		SC \cite{zelnik2005self}	&0.425	&0.183	&0.210	&0.123	&0.195	&0.091	&0.312	&0.136\\
		AC \cite{gowda1978agglomerative}	&0.371	&0.188	&0.222	&0.151	&0.209	&0.091	&0.329	&0.138\\
		NMF	\cite{cai2009locality} &0.420	&0.208	&0.275	&0.161	&0.187	&0.091	&0.305	&0.127\\
		PCA \cite{martinez01_pca}	&0.388	&0.198	&0.227	&0.173	&0.220	&0.090	&0.320	&0.139\\
		BIRCH \cite{Zhang1996}	&0.396	&0.202	&0.222	&0.147	&0.205	&0.092	&0.329	&-\\
		GMM	\cite{fraley03} &0.382	&0.193	&0.239	&0.169	&0.189	&0.091	&0.318	&0.132\\
		\midrule
		DEC	\cite{xie2016unsupervised} &0.534	&0.147	&0.257	&0.185	&0.232	&0.092	&0.421	&0.157\\
		IDEC \cite{guo2017improved}	&0.458	&0.141	&0.255	&0.192	&0.213	&0.094	&0.424	&0.160\\
		ASPC-DA \cite{guo2019adaptive}	&0.464	&0.073	&0.183	&0.079	&0.107	&0.067	&0.325	&-\\
		IDFD \cite{tao2021clustering}	&\textbf{0.595}	&0.456	&0.545	&0.628	&0.507	&0.306	&0.512	&0.424\\
		CC \cite{li2021contrastive}	&0.538	&0.480	&0.604	&0.622	&0.511	&0.358	&0.575	&0.426\\
		VTCC(Ours)	&0.572	&\textbf{0.553}	&\textbf{0.670}	&\textbf{0.716}	&\textbf{0.575}	&\textbf{0.393}	&\textbf{0.585}	&\textbf{0.455}\\
		\bottomrule
	\end{tabular}
\end{threeparttable}
\end{table*}

\begin{table*}[!t]\vskip 0.05in
	\centering
	\caption{The \textbf{ARI} scores of different image clustering methods on eight datasets.}\vskip -0.08in
	\label{tab-ARI}
\begin{threeparttable} \begin{tabular}{m{2cm}<{}|m{1.5cm}<{\centering}m{1.5cm}<{\centering}m{1.5cm}<{\centering}m{1.5cm}<{\centering}m{1.5cm}<{\centering}m{1.5cm}<{\centering}m{1.5cm}<{\centering}m{1.5cm}<{\centering}}
		\toprule
		Dataset &RSOD  &UC-Merced &SIRI-WHU &AID   &D0    &DTD   &Chaoyang &CIFAR-100\\
		\midrule
		$K$-means \cite{macqueen1967some}	&0.075	&0.065	&0.053	&0.051	&0.080	&0.016	&0.017	&0.028\\
		SC \cite{zelnik2005self}	&0.096	&0.038	&0.041	&0.029	&0.039	&0.012	&0.005	&0.022\\
		AC \cite{gowda1978agglomerative}	&0.071	&0.057	&0.057	&0.048	&0.080	&0.015	&0.010	&0.034\\
		NMF \cite{cai2009locality}	&0.052	&0.089	&0.118	&0.056	&0.068	&0.017	&0.002	&0.023\\
		PCA \cite{martinez01_pca}	&0.075	&0.064	&0.063	&0.054	&0.088	&0.015	&0.017	&0.029\\
		BIRCH \cite{Zhang1996}	&0.068	&0.066	&0.049	&0.046	&0.080	&0.016	&0.010	&-\\
		GMM	\cite{fraley03} &0.069	&0.062	&0.062	&0.053	&0.074	&0.015	&0.016	&0.028\\
		\midrule
		DEC	\cite{xie2016unsupervised} &0.325	&0.053	&0.083	&0.075	&0.105	&0.017	&0.006	&0.039\\
		IDEC \cite{guo2017improved}	&0.144	&0.042	&0.079	&0.073	&0.093	&0.017	&-	&0.042\\
		ASPC-DA \cite{guo2019adaptive}	&0.005	&0.002	&0.035	&0.014	&0.021	&0.005	&0.005	&-\\
		IDFD \cite{tao2021clustering}	&0.362	&0.354	&0.389	&0.547	&0.439	&0.174	&0.259	&0.276\\
		CC \cite{li2021contrastive}	&0.371	&0.356	&0.450	&0.550	&0.423	&0.205	&0.343	&0.267\\
		VTCC(Ours)	&\textbf{0.482}	&\textbf{0.453}	&\textbf{0.554}	&\textbf{0.622}	&\textbf{0.509}	&\textbf{0.233}	&\textbf{0.351}	&\textbf{0.281}\\
		\bottomrule
	\end{tabular}
\end{threeparttable}\vskip 0.1in
\end{table*}

\subsection{Comparisons with State-of-the-Art methods}
\label{sec:comparisons with state-of-the-art methods}

In this section, we evaluate the performance of the proposed VTCC method against twelve baseline clustering methods, including seven traditional clustering methods and five deep clustering methods. The seven traditional clustering methods include $K$-means \cite{macqueen1967some}, Spectral Clustering (SC) \cite{zelnik2005self}, Agglomerative Clustering (AC) \cite{gowda1978agglomerative}, Non-negative Matrix Factorization (NMF) \cite{cai2009locality}, Principle Component Analysis (PCA) \cite{martinez01_pca}, Balanced Iterative Reducing and Clustering using Hierarchies (BIRCH) \cite{Zhang1996}, Gaussian Mixture Model (GMM) \cite{fraley03}, whereas the five deep clustering methods include Deep Embedding Clustering (DEC) \cite{xie2016unsupervised}, Improved Deep Embedding Clustering (IDEC) \cite{guo2017improved}, Adaptive Self-Paced Deep Clustering with Data Augmentation (ASPC-DA) \cite{guo2019adaptive}, Instance Discrimination and Feature Decorrelation (IDFD) \cite{tao2021clustering}, and Contrastive Clustering (CC) \cite{li2021contrastive}. For NMF and PCA, the clustering results are obtained by conducting $K$-means on the dimensionally-reduced features. For the other baseline algorithms, the hyper-parameters will be set as suggested by their corresponding papers. For each test method, the number of clusters is set to the true number of classes on this dataset. The experimental results w.r.t. NMI, ACC, and ARI by different deep and non-deep clustering methods are reported in Tables ~\ref{tab-NMI}, \ref{tab-ACC}, and \ref{tab-ARI}, respectively.

In terms of NMI, as shown in Table~\ref{tab-NMI}, our VTCC method outperforms the baseline methods on all the eight datasets. Especially, on the RSOD dataset, VTCC achieves an NMI of 0.611, while the second best method (i.e., the CC method) only obtains an NMI of 0.457, where a relative gain of 33.7\% is observed. It is worth noting that the RSOD dataset is a small dataset with only 976 samples and that our VTCC model is trained from scratch. As for the other bigger datasets, our VTCC method also exhibits better or significantly better clustering performance than the baseline methods. In terms of ACC and ARI, as shown in Tables~\ref{tab-ACC} and ~\ref{tab-ARI}, similar advantages of VTCC can be seen in comparison with the other deep and non-deep clustering methods. The experimental results demonstrate the superiority of our VTCC method which jointly leverages Transformer encoder and contrastive learning for the clustering of complex images.

\subsection{Ablation Study}
\label{sec:ablation study}

In this section, we experimentally analyze the influence of different components in VTCC, including the convolutional stem, the ViT architecture, and the two contrastive projectors, on the RSOD and Chaoyang datasets.

\subsubsection{Influence of Convolutional Stem}
\label{sec:influence of convolutional stem}

This section conducts experiments to evaluate the influence of the convolutional stem, which is described in Section~\ref{sec:convolutional stem}. As shown in Table~\ref{tab:ablation_conv_stem}, using convolutional stem leads to an NMI score 0.611 on the RSOD dataset and 0.373 on the Chaoyang dataset, both of which are better than that of using the original patchify stem in ViT. Similar advantages in terms of ACC and ARI can also be observed, which confirm that the use of the convolutional stem is beneficial to the clustering performance of VTCC.

\begin{table}[!t]\scriptsize
	\caption{The influence of the convolutional stem.}\vskip -0.08in
	\label{tab:ablation_conv_stem}
	\renewcommand
	\arraystretch{1.25}
	\centering
	\scalebox{1}{
        \begin{threeparttable}
		\begin{tabular}{m{0.9cm}<{\centering}m{2.3cm}<{\centering}m{0.8cm}<{\centering}m{0.8cm}<{\centering}m{0.8cm}<{\centering}}
			\toprule
			Dataset &Split method  &NMI &ACC &ARI\\
			\midrule
			\multirow{2}*{RSOD}
			&Patchify Stem  &0.570 &0.565 &0.461\\
			&Convolutional Stem  &\textbf{0.611} &\textbf{0.572} &\textbf{0.482}\\
			\midrule
			\multirow{2}*{Chaoyang}
			&Patchify Stem  &0.365 &0.573 &0.336\\
			&Convolutional Stem  &\textbf{0.373} &\textbf{0.585} &\textbf{0.351}\\
			\bottomrule
	\end{tabular}
\end{threeparttable}}
\end{table}

\begin{table}[!t]\scriptsize
	\caption{The influence of ViT architecture.}\vskip -0.08in
	\label{tab:ablation_backbone}
	\renewcommand
	\arraystretch{1.25}
	\centering
	\scalebox{1}{
		\begin{threeparttable}
			\begin{tabular}{m{0.75cm}<{\centering}m{0.67cm}<{\centering}m{0.81cm}<{\centering}m{0.55cm}<{\centering}m{0.58cm}<{\centering} m{0.59cm}<{\centering}m{0.59cm}<{\centering}m{0.59cm}<{\centering}}
				\toprule
				Dataset &Model  &Dimension &\#Blocks &\#Heads &NMI &ACC &ARI\\
				\midrule
				\multirow{3}*{RSOD}
				&\mbox{ViT-Tiny} &192 &4 &12 &0.557 &0.563 &0.454\\
				&\mbox{ViT-Small} &384 &8 &12 &\textbf{0.611} &\textbf{0.572} &\textbf{0.482}\\
				&\mbox{ViT-Base} &768 &12 &12 &0.480 &0.554 &0.393\\
				\midrule
				\multirow{3}*{Chaoyang}
				&\mbox{ViT-Tiny} &192 &4 &12 &0.340 &0.580 &0.328\\
				&\mbox{ViT-Small} &384 &8 &12 &\textbf{0.373} &\textbf{0.585} &\textbf{0.351}\\
				&\mbox{ViT-Base} &768 &12 &12 &0.362 &0.571 &0.332\\
				\bottomrule
			\end{tabular}
	\end{threeparttable}}
\end{table}

\begin{table}[!t]\scriptsize
	\caption{The influence of two contrastive projectors.}\vskip -0.08in
	\label{tab:ablation_head}
	\renewcommand
	\arraystretch{1.25}
	\centering
	\scalebox{1}{
		\begin{threeparttable}	\begin{tabular}{m{0.75cm}<{\centering}m{4cm}<{\centering}m{0.59cm}<{\centering}m{0.59cm}<{\centering}m{0.59cm}<{\centering}}
				\toprule
				Dataset &Contrastive projectors  &NMI &ACC &ARI\\
				\midrule
				\multirow{3}*{RSOD}
				&Instance projector only + $K$-means &0.465 &\textbf{0.628} &0.362\\
				&Cluster projector only  &0.529 &0.618 &0.425\\
				&Instance projector + Cluster projector  &\textbf{0.611} &0.572 &\textbf{0.482}\\
				\midrule
				\multirow{3}*{Chaoyang}
				&Instance projector only + $K$-means &0.343 &0.478 &0.240\\
				&Cluster projector  only&0.310 &0.546 &0.279\\
				&Instance projector + Cluster projector  &\textbf{0.373} &\textbf{0.585} &\textbf{0.351}\\
				\bottomrule
			\end{tabular}
	\end{threeparttable}}\vskip 0.1in
\end{table}

\subsubsection{Influence of Vision Transformer Architecture}
\label{sec:influence of vision transformer architecture}

To evaluate the influence of the ViT architecture, we compare the performances of VTCC with different scales of Transformer encoder architectures, including ViT-Tiny \cite{touvron2021training}, ViT-Small \cite{touvron2021training}, and ViT-Base \cite{touvron2021training}, and report the clustering results in Table~\ref{tab:ablation_backbone}. As shown in Table~\ref{tab:ablation_backbone}, on the small dataset of RSOD, ViT-Small and ViT-Tiny outperform ViT-Base, probably due to the fact that ViT-Base is the largest model (among the three) and may lead to over-fitting on the small dataset. Empirically, the ViT-Small is a suitable choice with sufficient representation learning capability while maintaining high efficiency.

\subsubsection{Influence of Contrastive Projectors}
\label{sec:influence of contrastive head}

This section conducts experiments to evaluate the influence of the two contrastive projectors in VTCC. When testing the performance of the instance projector, the cluster projector is removed and the $K$-means clustering is performed on the representation learned by the instance projector. As shown in Table~\ref{tab:ablation_head}, on the RSOD dataset, only using the instance projector leads to an NMI of 0.465, whereas only using the cluster projector leads to an NMI of 0.529, both of which are lower than the NMI of jointly using two projectors, that is, 0.611. Although only using the instance projector can lead to a better ACC score than using both projectors, yet using both projectors yields the best performance in terms of the other two metrics on the RSOD dataset, and  in terms of all the three metrics on the Chaoyang dataset.

\subsection{t-SNE Visualization}
\label{sec:visual analysis}
\begin{figure*}[!t]\vskip 0.1in
	\centering
	\subfigure[0 epoch (NMI=0.231)]{
		\includegraphics[width=0.234\textwidth]{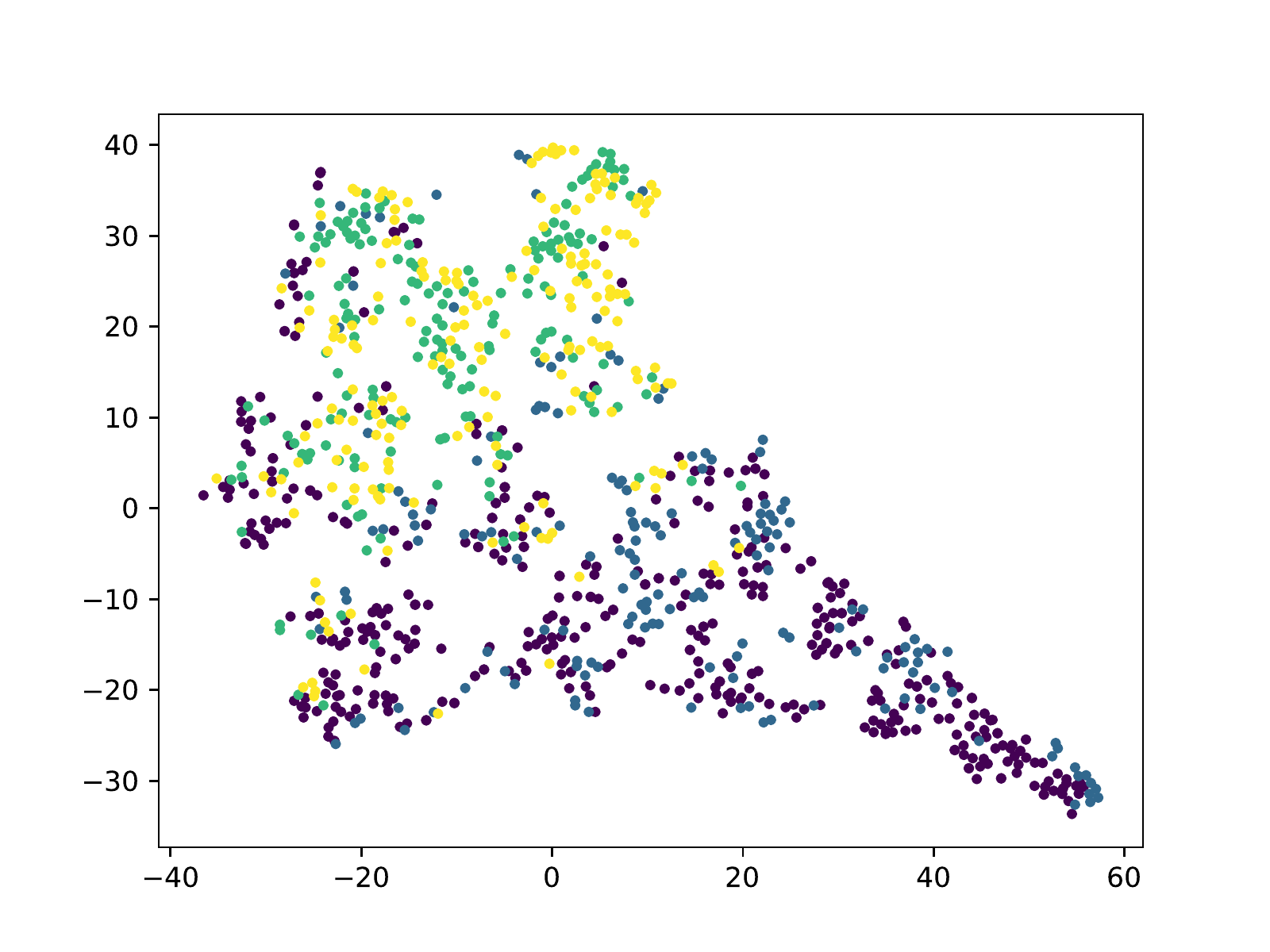}}
	\subfigure[50 epoch (NMI=0.454)]{
		\includegraphics[width=0.234\textwidth]{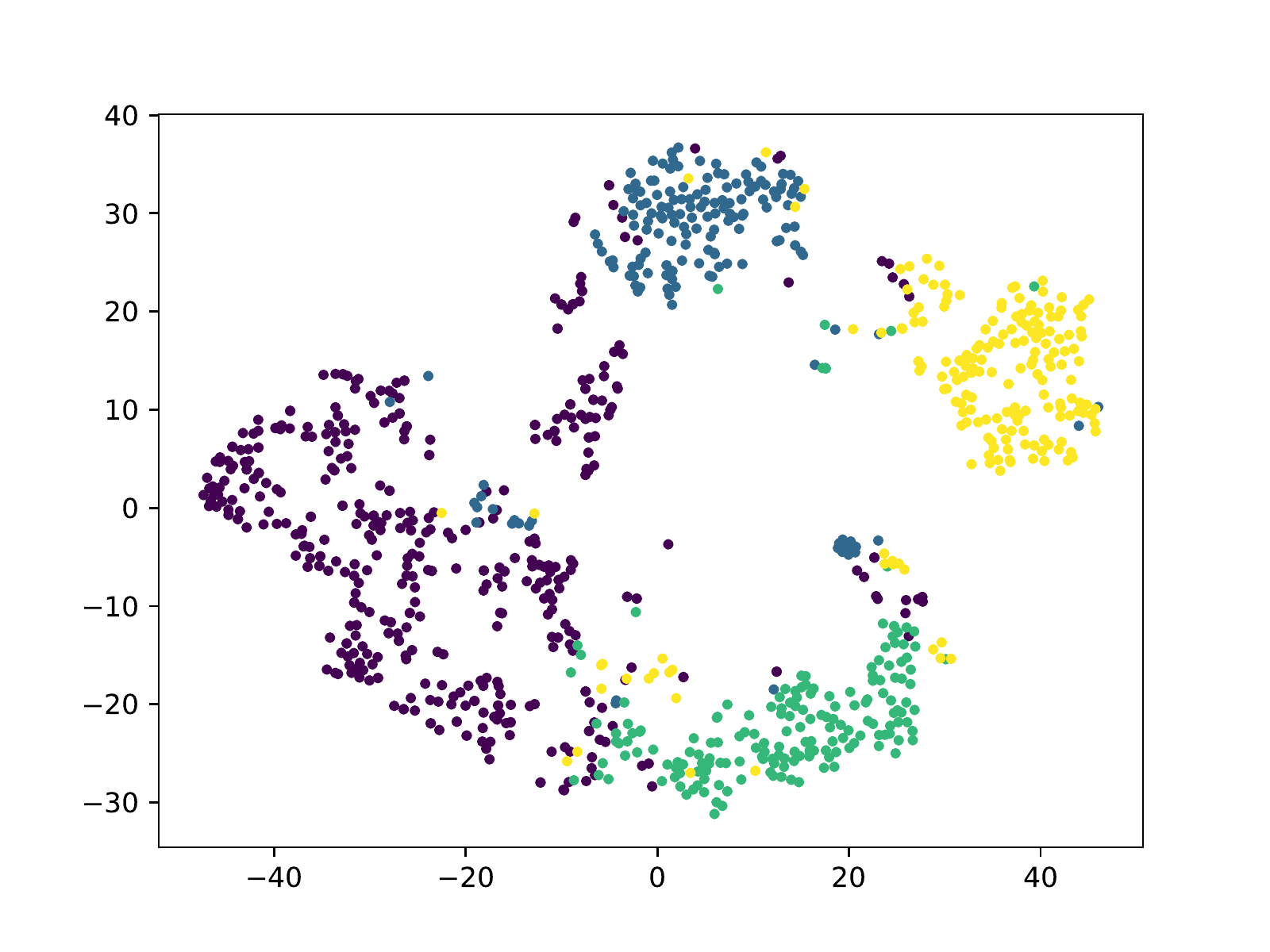}}
	\subfigure[100 epoch (NMI=0.506)]{
		\includegraphics[width=0.234\textwidth]{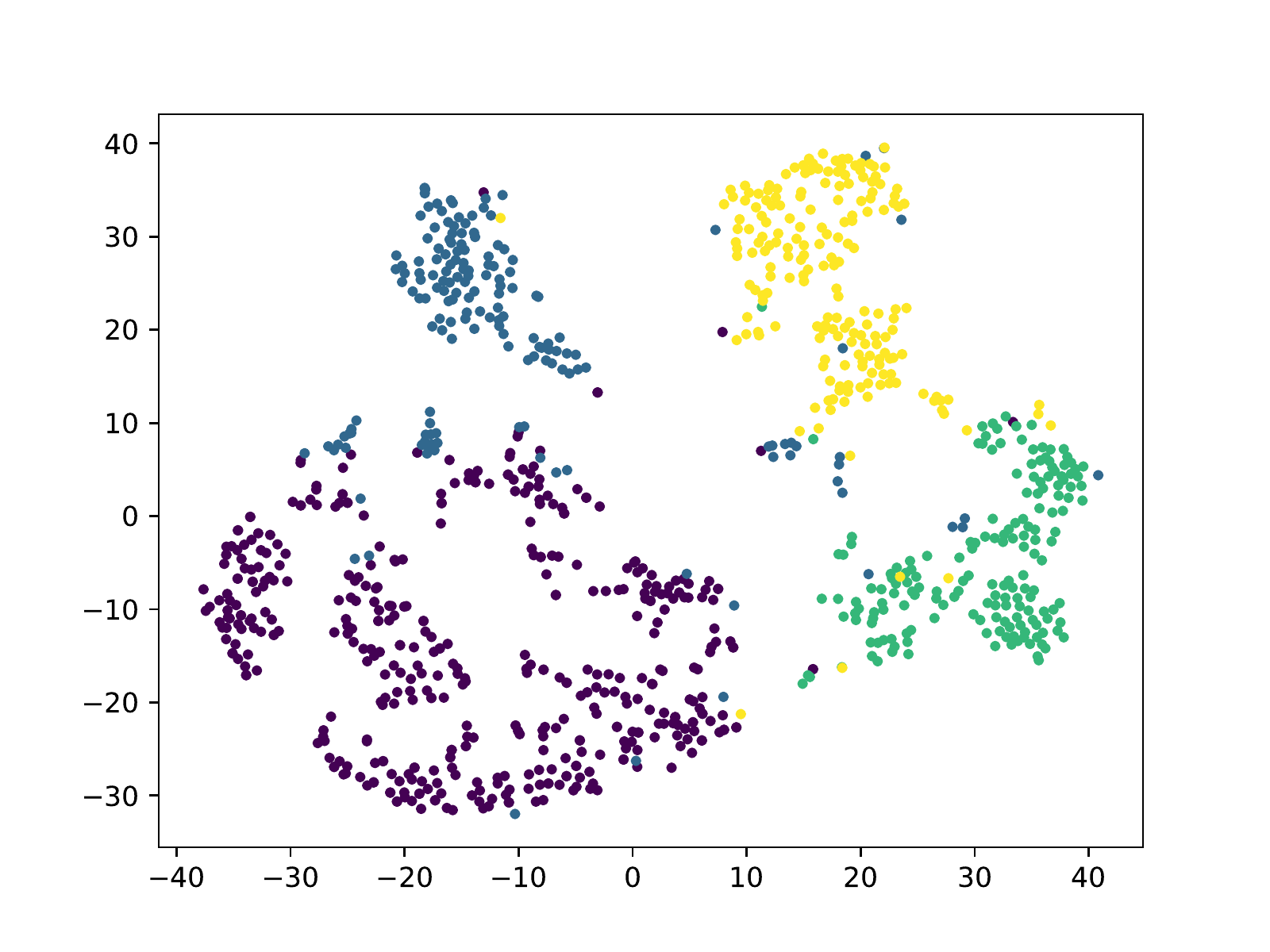}}
	\subfigure[500 epoch (NMI=0.611)]{
		\includegraphics[width=0.234\textwidth]{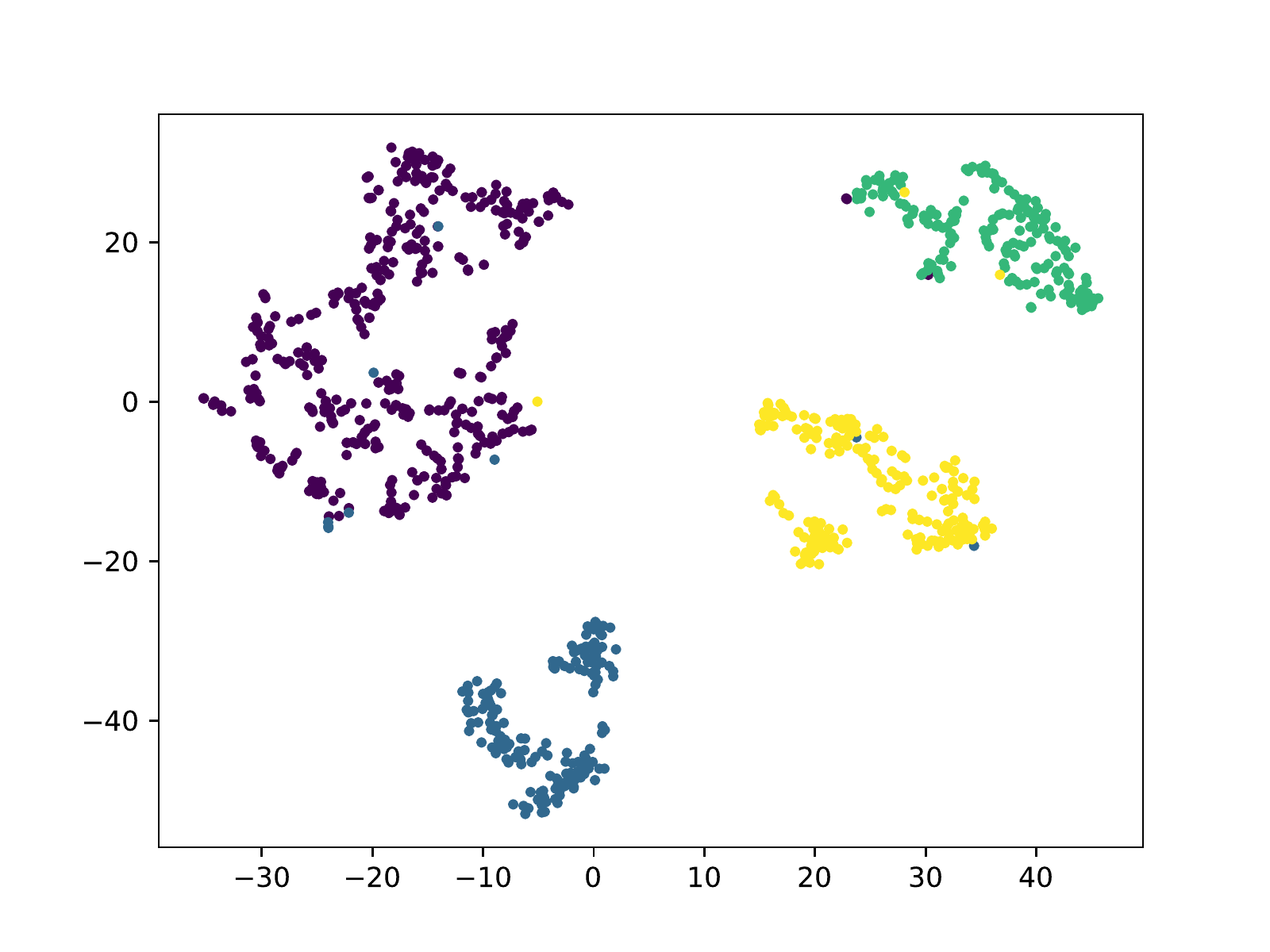}}
	\caption{The t-SNE visualization of VTCC on the RSOD dataset.}\vskip 0.05 in
	\label{fig:t-SNE}
\end{figure*}

Our VTCC method is capable of learning discriminative features by jointly exploiting the Transformer and the contrastive learning. In Fig.~\ref{fig:t-SNE}, we use the t-SNE \cite{van2008visualizing} visualization to show the changes of the representation quality with varying number of epochs. As shown in Fig.~\ref{fig:t-SNE}, the quality of the feature representations learned by VTCC improves as the number of the training epochs goes from 0 to 500, where clear separation between clusters can be observed.

\section{Conclusion}
\label{sec:conclusion}

This paper presents a novel deep clustering approach termed VTCC, which for the first time unifies the Transformer and the contrastive learning for the image clustering task. In particular, we utilize a ViT encoder architecture as the backbone to extract the feature representations of two augmented views, where a convolutional stem is used to split each augmented sample into a sequence of patches for the Transformer encoder. Two contrastive projectors are incorporated to simultaneously enforce the instance-level contrastive learning and the cluster-level contrastive learning (for global clustering structure learning). Extensive experiments are conducted on eight real-world image datasets, which have demonstrated the superiority of VTCC over the state-of-the-art deep clustering approaches. Notably, the experimental results reveal the benefits of the joint modeling of Transformer and contrastive learning for unsupervised image clustering, especially for images with complex structures. Starting from the VTCC approach, more Transformer architectures and contrastive learning strategies can be explored for the unsupervised image clustering task as well as other unsupervised learning tasks.

\ifCLASSOPTIONcaptionsoff
\newpage
\fi

\bibliographystyle{IEEEtran}

\bibliography{VTCC}

\end{document}